\documentclass[letterpaper]{article} 
\usepackage{aaai25}  
\usepackage{times}  
\usepackage{helvet}  
\usepackage{courier}  
\usepackage[hyphens]{url}  
\usepackage{graphicx} 
\usepackage{natbib}  
\usepackage{caption} 
\usepackage{algorithmic}
\usepackage[ruled,vlined]{algorithm2e}
\usepackage[frozencache=true,cachedir=minted-cache]{minted} 

\usepackage{amssymb}
\usepackage{booktabs} 
\usepackage{makecell}
\usepackage{amsmath}
\usepackage{amsfonts}
\usepackage{multirow}
\usepackage{bbding}
\usepackage{color,xcolor}
\usepackage{array}
\usepackage{threeparttable}
\usepackage{siunitx}
\usepackage{pifont}

\usepackage{newfloat}
\usepackage{listings}
\DeclareCaptionStyle{ruled}{labelfont=normalfont,labelsep=colon,strut=off} 
\floatstyle{ruled}
\newfloat{listing}{tb}{lst}{}
\floatname{listing}{Listing}
\title{Gradient Alignment Improves Test-Time Adaptation for Medical Image Segmentation}
\author {
    Ziyang Chen\textsuperscript{\rm 1}, 
    Yiwen Ye\textsuperscript{\rm 1},
    Yongsheng Pan\textsuperscript{\rm 1 $\dag$}, 
    Yong Xia\textsuperscript{\rm 1,2,3 $\dag$}
}

\affiliations {
    \textsuperscript{\rm 1} National Engineering Laboratory for Integrated Aero-Space-Ground-Ocean Big Data Application Technology, School of Computer Science and Engineering, Northwestern Polytechnical University, Xi'an, China\\
    \textsuperscript{\rm 2} Research \& Development Institute of Northwestern Polytechnical University in Shenzhen, Shenzhen, China\\
    \textsuperscript{\rm 3} Ningbo Institute of Northwestern Polytechnical University, Ningbo, China\\
    \{zychen, ywye\}@mail.nwpu.edu.cn, \{yspan, yxia\}@nwpu.edu.cn
}

\usepackage{bibentry}

\begin{document}

\maketitle
\footnotetext{$^\dag$Corresponding author.}

\begin{abstract}
Although recent years have witnessed significant advancements in medical image segmentation, the pervasive issue of domain shift among medical images from diverse centres hinders the effective deployment of pre-trained models. 
Many Test-time Adaptation (TTA) methods have been proposed to address this issue by fine-tuning pre-trained models with test data during inference. 
These methods, however, often suffer from less-satisfactory optimization due to suboptimal optimization direction (dictated by the gradient) and fixed step-size (predicated on the learning rate).
In this paper, we propose the \textbf{Gr}adient \textbf{a}lignment-based \textbf{T}est-time \textbf{a}daptation (GraTa) method to improve both the gradient direction and learning rate in the optimization procedure.
Unlike conventional TTA methods, which primarily optimize the pseudo gradient derived from a self-supervised objective, our method incorporates an auxiliary gradient with the pseudo one to facilitate gradient alignment.
Such gradient alignment enables the model to excavate the similarities between different gradients and correct the gradient direction to approximate the empirical gradient related to the current segmentation task.
Additionally, we design a dynamic learning rate based on the cosine similarity between the pseudo and auxiliary gradients, thereby empowering the adaptive fine-tuning of pre-trained models on diverse test data.
Extensive experiments establish the effectiveness of the proposed gradient alignment and dynamic learning rate and substantiate the superiority of our GraTa method over other state-of-the-art TTA methods on a benchmark medical image segmentation task.
\end{abstract}

%
\begin{links}
    \link{Code}{https://github.com/Chen-Ziyang/GraTa}
\end{links}


\section{Introduction}
Medical image segmentation assumes a pivotal role in computer-aided diagnosis, offering precise delineation of specific anatomical structures.
Over the past years, considerable research efforts~\cite{SegReview_1,SegReview_2,SegReview_3,wang2022stepwise,tang2023duat,xu2024polyp} have been devoted to medical image segmentation utilizing deep-learning techniques, resulting in significant progress.
However, domain shift, primarily caused by variations in scanners, imaging protocols, and operators~\cite{domain_shift_1,domain_shift_2}, poses challenges for these models pre-trained on the labeled dataset (source domain) to generalize across different medical centers (target domain).

\begin{figure*}[!tb]
	\centering
	\includegraphics[width=0.7\textwidth]{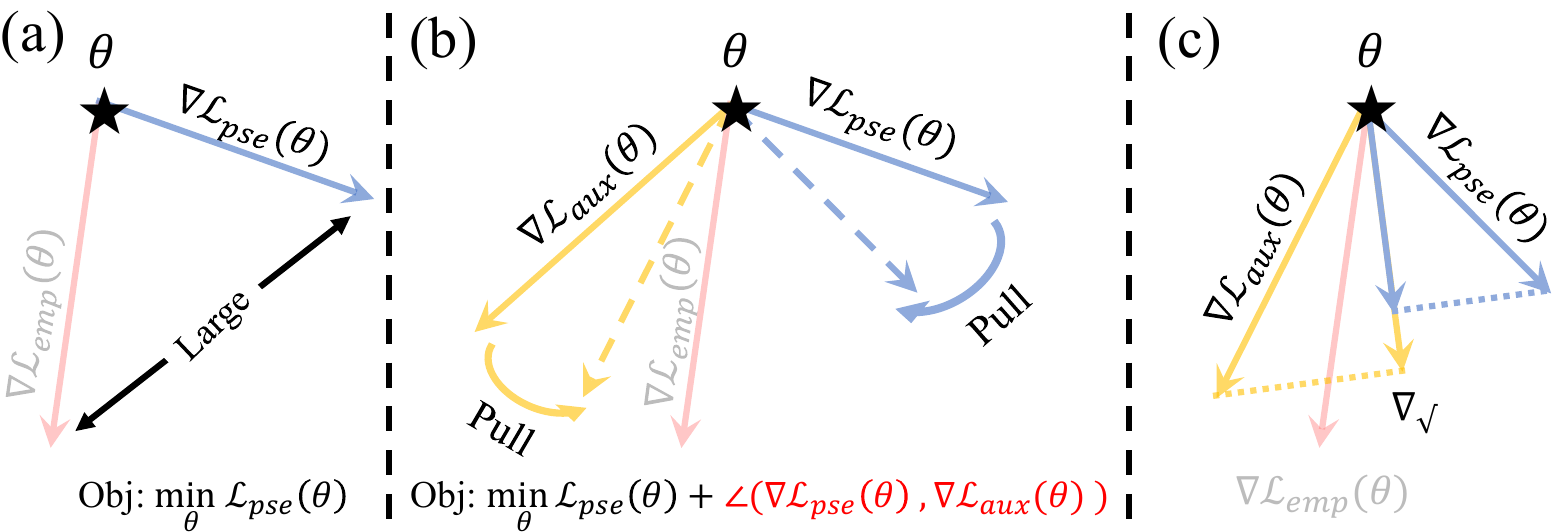}
	\caption{Illustration of our motivation. 
        (a) We display a pseudo gradient $\nabla\mathcal{L}_{pse}(\theta)$, which is primarily utilized for optimization but may diverge far from the empirical gradient $\nabla\mathcal{L}_{emp}(\theta)$, tailored to the specific task (segmentation in this study).
        Existing methods typically optimize $\nabla\mathcal{L}_{pse}(\theta)$ in a straightforward manner.
        (b) Our GraTa introduces an auxiliary gradient $\nabla\mathcal{L}_{aux}(\theta)$ to minimize the angle between $\nabla\mathcal{L}_{pse}(\theta)$ and $\nabla\mathcal{L}_{aux}(\theta)$, resulting in a novel auxiliary objective, \emph{i.e.}, gradient alignment.
        (c) Although achieving complete alignment is challenging due to the different objectives of these two gradients, the model can learn to align their task-relevant components $\nabla_\checkmark$ through this auxiliary objective, approximating $\nabla\mathcal{L}_{emp}(\theta)$ and facilitating effective fine-tuning. 
        $\angle$ denotes the angle.
        `Obj': Abbreviation of `Objective'.
 }
\label{fig-motivation}
\end{figure*}

Test-time adaptation (TTA) has emerged as a prospective paradigm to mitigate domain shift with minimal data demand, relying only on test data during inference.
Several studies on TTA attempt to alleviate the domain shift by modifying the statistics stored in batch normalization (BN) layers~\cite{Dynamically,BN,DUA,DomainAdaptor,MedBN}.
A typical example is DUA~\cite{DUA}, which incrementally adjusts the statistics within BN layers from the source to the target domain to enhance feature representations.
Although these methods improve adaptation, their performance potential is limited since the model's parameters remain fixed.
The mainstream TTA methods focus on designing self-supervised objectives (\emph{e.g.}, entropy minimization, consistency constraint, and rotation prediction) to fine-tune the pre-trained models to boost their generalization performance on the target domain~\cite{TENT,TTT,DomainAdaptor,SAR,TIPI,TeST,DLTTA,BatesonTTA,DeTTA}.
For instance, TENT~\cite{TENT} introduces an entropy minimization objective to TTA to optimize the affine-transformation parameters of pre-trained models, and TTT~\cite{TTT} updates the model parameters by predicting the rotation angle of test data in a self-supervised manner.

However, all these methods, based on gradient descent, overlook two critical elements in the optimization procedure: the direction and the step-size.
The $i$th optimization procedure of the pre-trained model with parameters $\theta$ can be formulated as $\theta_{i+1}\leftarrow \theta_{i} - \eta \nabla\mathcal{L}_{pse}(\theta_{i})$, where $\eta$ is the learning rate, and $\nabla\mathcal{L}_{pse}$ denotes the gradient produced by the self-supervised objective function, called pseudo gradient. 
Obviously, the optimization direction is determined by $\nabla\mathcal{L}_{pse}(\theta_{i})$, and the optimization step-size depends on $\eta$.

In Figure~\ref{fig-motivation}(a), we displayed two gradients: an empirical gradient, customized for the specific task (segmentation in this study) using provided labels and not encountered in the TTA setup; and a pseudo gradient, primarily employed for fine-tuning pre-trained models in existing TTA methods.
Ideally, the pseudo gradient should exhibit a direction similar to the empirical gradient.
Unfortunately, due to the lack of reliable supervision, their directions may differ significantly, posing challenges to model optimization.
Current TTA methods simply optimize the pseudo gradient, failing to address this issue.
Moreover, concerning the learning rate, almost all these methods employ a fixed value, thereby restricting the pre-trained models from being adaptively fine-tuned on diverse test data.

In this paper, we focus on optimizing both the optimization direction and the learning rate.
We propose a novel TTA method, namely \textbf{Gr}adient \textbf{a}lignment-based \textbf{T}est-time \textbf{a}daptation (GraTa). 
GraTa aims to improve the gradient direction through gradient alignment and enables pre-trained models to adapt to test data using a dynamic learning rate.
As shown in Figure~\ref{fig-motivation}(b), we incorporate a new auxiliary objective, \emph{i.e.}, gradient alignment, by introducing an auxiliary gradient to the pseudo one.
Specifically, we employ the consistency loss to produce the pseudo gradient, while leveraging the gradient derived from the entropy loss as the auxiliary one.
The entropy loss is computed on the original test data. The consistency loss is conducted on the weak and strong augmentation variants of the test data.
Through alignment, the model is capable of excavating the similarities between distinct gradients, especially the components relevant to the current segmentation task, and approximating the empirical gradient (see Figure~\ref{fig-motivation}(c)).
Furthermore, we also present a dynamic learning rate, which is inversely proportional to the angle between these two gradients, to adaptively fine-tune the pre-trained models.
A larger angle implies greater conflict in the optimization of the two gradients, thus requiring a smaller learning rate, and vice versa.
Extensive experiments demonstrate that our GraTa achieves a superior performance over other state-of-the-art TTA methods.

Our contributions are three-fold:
(1) we rethink the optimization procedure in existing TTA methods and propose GraTa to improve both the optimization direction and step-size; 
(2) we align the pseudo and auxiliary gradients, aimed at distinct objectives, to reduce the divergence of their task-specific components, thereby improving the optimization direction; and 
(3) we design a variable learning rate considering the angle between distinct gradients, aiding in the dynamical determination of the optimization step-size for adaptive fine-tuning.


\section{Related Work}
\label{sec:work}

\subsection{Test-time Adaptation}
Test-time Adaptation (TTA) addresses domain shift by adapting the pre-trained source model to the distribution of the target domain using test data during inference~\cite{TTA_survey1}. TTA can be employed in a source-free and online manner, allowing the model to adapt without accessing the source data.
The mainstream TTA methods are based on fine-tuning the model by constructing self-supervised auxiliary tasks to guide the model in training on the test data, enhancing its performance and adaptability~\cite{TENT,TTT,MEMO,T3A,TIPI,SAR,DLTTA,DomainAdaptor,DeTTA}. 
\cite{TENT} proposed a test-time entropy minimization scheme to decrease the entropy of model predictions by fine-tuning the affine parameters within batch normalization layers.
\cite{DomainAdaptor} extended the vanilla entropy minimization loss to a generalized one to better utilize the information in the test data.
\cite{SAR} filtered and removed partial noise with large uncertainty in the supervised information and employed sharpness-aware minimization to optimize the parameters of the model toward a flat minimum.
These methods demonstrate effective adaptation when the supervised information is reliable but may lead to unexpected performance degradation when the supervised information is unreliable.
In this paper, our purpose is to improve the optimization direction and step-size by gradient alignment and dynamic learning rate, respectively.
After aligning the gradients, we can obtain more reliable supervised information, thereby facilitating more effective training of the source model.


\subsection{Gradient Alignment}
In deep learning, the objective function serves as a measure to quantify the discrepancy between a model's prediction and the expected output. Objective functions with similar gradients generally indicate similar objectives and contribute to robust training, which is particularly important in multi-task learning. 
A typical approach in multi-task learning is PCGrad~\cite{GradientSurgery}, which is developed to mitigate gradient conflicts between task gradients via replacing the gradient by its projection onto the normal plane of another gradient.
In contrast, our optimization objective does not involve resolving conflicts between multiple gradients, as it does not directly optimize multiple gradients simultaneously. Instead, the goal of our GraTa is to improve the direction of pseudo gradient, thereby enhancing the model's fine-tuning process.
Recently, many methods seek to align the gradients of multiple objective functions to improve training~\cite{FGDA,IGA,Joo,GradientCVPR,GradientAAAI,GradientICCV}.
\cite{FGDA} devised a gradient discriminator to align the gradients from the source and target domains to improve the similarity between two distributions for better adversarial domain adaptation.
\cite{IGA} proposed an algorithm for federated learning that induces the implicit regularization to promote the alignment of gradients across different clients.
\cite{Joo} presented to combine both L2 and cosine distance-based criteria as regularization terms, leveraging their strengths in aligning the local gradient to produce more robust interpretations.
Different from these methods, we implicitly perform gradient alignment to capture the similarity between gradients of two distinct objective functions to improve the gradient direction for test-time adaptation.

\subsection{Dynamic Learning Rate}
The learning rate is a critical hyperparameter in deep learning, as it determines the step-size to update the model. An excessively large learning rate may hinder the model's convergence, while an excessively small learning rate may make the model trapped in local optima. Extensive research has been conducted to dynamically adjust the learning rate to enhance model training~\cite{DLTTA,DynamicCVPR,DynamicNIPS,DynamicICLR,DynamicTNN}.
\cite{DLTTA} presented a dynamic learning rate by constructing a memory bank and calculating the cosine similarity scores between the feature of the current test image and previous features from the memory bank.
\cite{DynamicCVPR} devised the dynamic weighted learning to dynamically weights the learning losses of alignment and discriminability to avoid excessive alignment learning and excessive discriminant learning. 
\cite{DynamicICLR} proposed a graph-network-based scheduler to learn a specific scheduling mechanism without restrictions and control the learning rate via reinforcement learning.
In this paper, our focus is on enhancing the optimization direction by aligning two distinct gradients. 
The cosine similarity between these two gradients can be employed as a natural metric to measure the degree of alignment, allowing our GraTa to adjust the learning rate dynamically.


\section{Method}
\subsection{Problem Definition}
\begin{figure*}[!tb]
	\centering
	\includegraphics[width=0.75\textwidth]{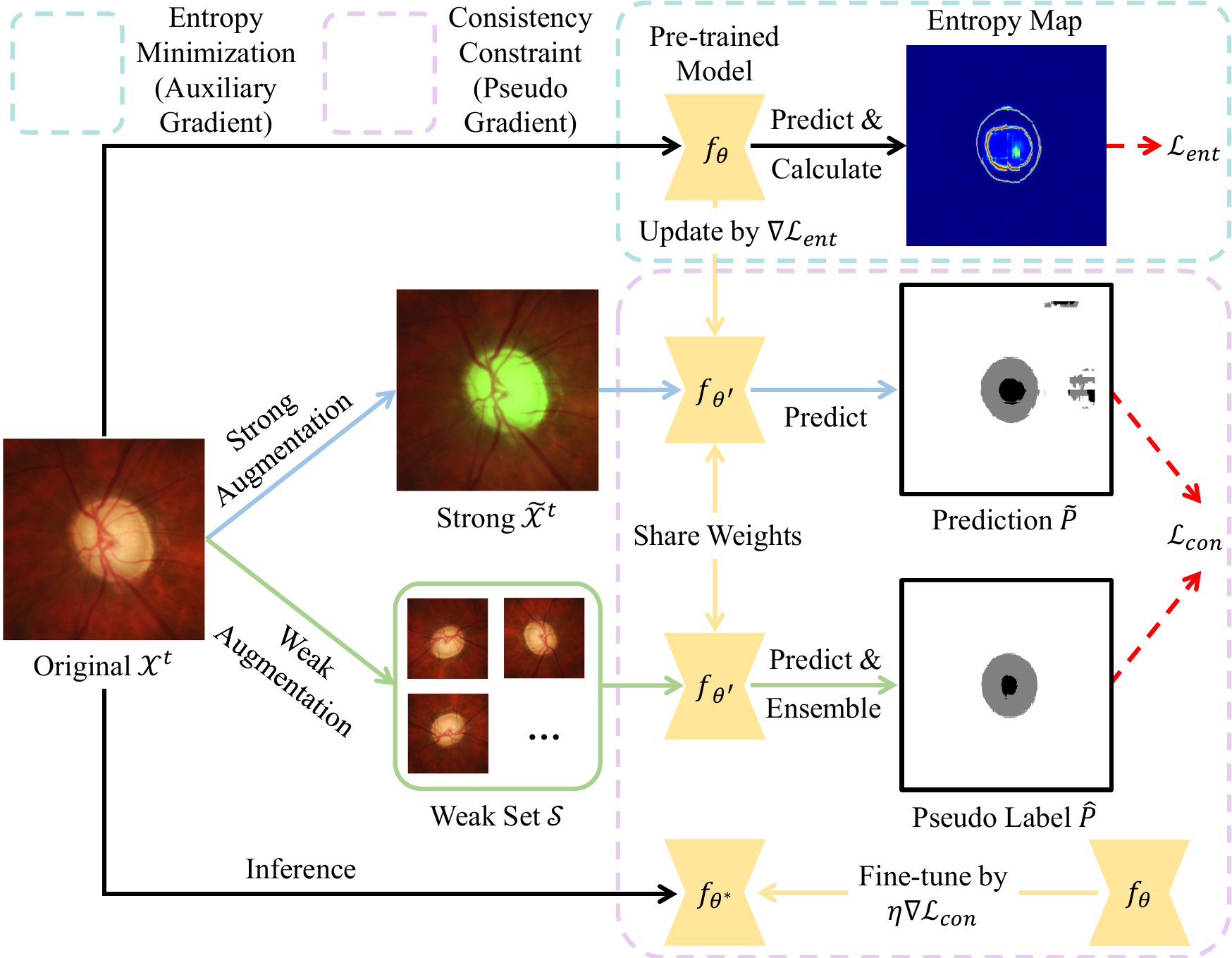}
	\caption{Overview of our GraTa. For each test image, we first calculate the entropy loss on its prediction to update the pre-trained model $f_{\theta}$ by $\theta^{'}=\theta-\nabla\mathcal{L}_{ent}$. Then we perform weak and strong augmentation on the original test image to obtain a strong augmentation variant and a set of weak augmentation variants and calculate the consistency loss on their predictions produced by $f_{\theta^{'}}$. Finally, the consistency loss is utilized to fine-tune the model by $\theta^{*}\leftarrow \theta-\eta\nabla\mathcal{L}_{con}$, and the test image is fed into $f_{\theta^{*}}$ for inference.
 }
\label{fig-method}
\end{figure*}

Let the labeled source domain dataset and unlabeled testing target dataset be 
$\mathcal{D}^{s} = \left \{ \mathcal{X}_{i}^{s}, \mathcal{Y}_{i}^{s} \right \}_{i=1}^{N^s}$ and $\mathcal{D}^{t} = \left \{ \mathcal{X}_{i}^{t}\right \}_{i=1}^{N^t}$, respectively, where $\mathcal{X}_i^{*}\in \mathbb{R}^{H \times W \times C}$ is the $i$-th image with C channels and size of $H\times W$, and $\mathcal{Y}_i^{*}$ is its label. 
Our goal is to fine-tune the model $f_\theta:\mathcal{X} \rightarrow \mathcal{Y}$ pre-trained on $\mathcal{D}^{s}$ using $\mathcal{X}_{i}^{t}$ to adapt to $\mathcal{D}^{t}$.
An overview of our GraTa is illustrated in Figure~\ref{fig-method}.
Now we delve into the details.



\subsection{The Objective of GraTa}
In this study, we introduce the entropy loss $\mathcal{L}_{ent}$ as the auxiliary objective function to align with the consistency loss $\mathcal{L}_{con}$.
Our GraTa aims at minimizing not only $\mathcal{L}_{con}$ but also the angle between gradients derived from $\mathcal{L}_{con}$ and $\mathcal{L}_{ent}$ for gradient alignment.
$\mathcal{L}_{ent}$ is calculated on the prediction $P_i$ of original test data $\mathcal{X}^t_i$ as follows:
\begin{equation}
\begin{aligned}
\mathcal{L}_{ent}(\theta;\mathcal{X}^t_i) = -P_i\log{P_i}, P_i=Sigmoid(f_\theta(\mathcal{X}^t_i)).
\end{aligned}
\end{equation}
$\mathcal{L}_{con}$ is conducted on the set of weak augmentation variants $\mathcal{S}_i=\{\hat{\mathcal{X}}^t_{ij} | \hat{\mathcal{X}}^t_{ij}=Aug^j_{w}(\mathcal{X}_i^t)\}_{j=1}^{6}$ and the strong augmentation variant $\tilde{\mathcal{X}}_i^t=Aug_{s}(\mathcal{X}_i^t)$ as follows:
\begin{equation}
\begin{aligned}
&\mathcal{L}_{con}(\theta;\mathcal{X}^t_i) = l_{ce}(\tilde{P_i},\hat{P_i}),
\tilde{P_i}=Sigmoid(f_\theta(\tilde{\mathcal{X}}^t_i)), \\
&\hat{P_i}=\frac{1}{j} \sum_{j=1}^{6} Sigmoid(f_\theta(\hat{\mathcal{X}}^t_{ij})),
\end{aligned}
\end{equation}
where $l_{ce}$ denotes the cross-entropy loss.
The weak augmentation strategy $Aug_{w}$ consists of six augmentation techniques: identity mapping, horizontal flipping, vertical flipping, rotation $90^\circ$, $180^\circ$, and $270^\circ$. We sequentially apply each $Aug^j_{w}$ to $\mathcal{X}^t_{i}$ to obtain $\mathcal{S}$.
The strong augmentation strategy $Aug_{s}$ incorporates brightness adjustment, contrast adjustment, Gamma transformation, Gaussian noise, and Gaussian blur.
$\mathcal{X}^t_{i}$ is transformed to $\tilde{\mathcal{X}}_i^t$ utilizing all of these techniques.

Then the overall objective can be formulated as
\begin{equation}
\begin{aligned}
\min_{\theta} \mathcal{L}_{con} (\theta;\mathcal{X}^t_i)+\angle (\nabla\mathcal{L}_{con} (\theta;\mathcal{X}^t_i),\nabla\mathcal{L}_{ent} (\theta;\mathcal{X}^t_i)),
\end{aligned}
\label{obj1}
\end{equation}
where $\angle$ denotes the angle.
However, the angle is non-differentiable and cannot be directly minimized, and a common solution is to maximize the inner product of these two gradients~\cite{SAGM}, \emph{i.e.}, $\nabla\mathcal{L}_{con} (\theta;\mathcal{X}^t_i) \cdot \nabla\mathcal{L}_{ent} (\theta;\mathcal{X}^t_i)$.
Unfortunately, calculating the inner product explicitly will introduce the calculation of the Hessian matrix during back-propagation, which is computationally complex and unstable.
Inspired by~\cite{MLDG}, we rewrite the objective to implicitly maximize the inner product as follows:
\begin{equation}
\begin{aligned}
\min_{\theta} \mathcal{L}_{con} (\theta^{'};\mathcal{X}^t_i),
\end{aligned}
\label{obj2}
\end{equation}
where $\theta^{'}=\theta-\nabla\mathcal{L}_{ent}(\theta;\mathcal{X}^t_i)$.

\noindent{\textbf{Proof.}}
For Eq.~\ref{obj2}, we perform the first-order Taylor expansion around $\theta$ as follows:
\begin{equation}
\begin{aligned}
& \min_\theta \mathcal{L}_{con} (\theta-\nabla\mathcal{L}_{ent}(\theta;\mathcal{X}^t_i);\mathcal{X}^t_i) \\
= & \min_\theta \mathcal{L}_{con} (\theta;\mathcal{X}^t_i)-\nabla\mathcal{L}_{con}(\theta;\mathcal{X}^t_i) \cdot \nabla\mathcal{L}_{ent}(\theta;\mathcal{X}^t_i) + O \\
\approx & \min_\theta \mathcal{L}_{con} (\theta;\mathcal{X}^t_i)-\nabla\mathcal{L}_{con}(\theta;\mathcal{X}^t_i) \cdot \nabla\mathcal{L}_{ent}(\theta;\mathcal{X}^t_i),
\end{aligned}
\label{derivation}
\end{equation}
where $O$ denotes the remainder term, which can be omitted. Then, the objective in Eq.~\ref{obj2} can be rewritten as:
\begin{equation}
\begin{aligned}
\min_{\theta} \mathcal{L}_{con} (\theta;\mathcal{X}^t_i) -\nabla\mathcal{L}_{con}(\theta;\mathcal{X}^t_i) \cdot \nabla\mathcal{L}_{ent}(\theta;\mathcal{X}^t_i).
\end{aligned}
\label{new_obj2}
\end{equation}
Eq.~(\ref{new_obj2}) reveals that our objective is to minimize $\mathcal{L}_{con} (\theta;\mathcal{X}^t_i)$ and maximize the inner product of these two gradients. 
As $\nabla\mathcal{L}_{con} (\theta;\mathcal{X}^t_i)$ and $\nabla\mathcal{L}_{ent} (\theta;\mathcal{X}^t_i)$ become more closely aligned, their inner product increases, reaching the maximum when they are completely aligned.


\SetKwInOut{Require}{Require}
\SetKwInOut{Initialize}{Initialize}
\begin{algorithm}[!htb]
\caption{The Algorithm of GraTa.}
\small
\Initialize{Pre-trained source model $f_{\theta_0}$ and scaling factor $\beta$.}
\KwIn{For each time step $i$ (starting from 0), current test image $\mathcal{X}_{i}^{t}$.}
\begin{algorithmic}

\STATE $\triangleright$ \textbf{Fine-tune the model:} 
\STATE 1: Calculate $\mathcal{L}_{ent}(\theta_i;\mathcal{X}^t_i)$ by Eq. (1) and perform back-propagation
\STATE 2: Update the model by $\theta_i^{'}=\theta_i-\nabla\mathcal{L}_{ent}(\theta_i;\mathcal{X}^t_i)$
\STATE 3: Calculate $\mathcal{L}_{con}(\theta_i^{'};\mathcal{X}^t_i)$ by Eq. (2) and perform back-propagation
\STATE 4: Obtain $\eta$ utilizing $\beta$ through Eq. (5)
\STATE 5: Fine-tune the model by $\theta_{i+1} \leftarrow \theta_i-\eta 
\nabla\mathcal{L}_{con}(\theta_i^{'};\mathcal{X}^t_i)$

\STATE $\triangleright$ \textbf{Inference:}
\STATE 6: Forward $\mathbb{P}_i=f_{\theta_{i+1}}(\mathcal{X}_{i}^{t})$

\end{algorithmic}
\KwOut{Adapted prediction $\mathbb{P}_i$}
\label{algorithm1}
\end{algorithm}

\subsection{Dynamic Learning Rate}
We design a dynamic learning rate $\eta$ for adaptive fine-tuning. 
$\eta$ is determined based on the cosine similarity between $\nabla\mathcal{L}_{con}(\theta^{'};\mathcal{X}^t_i)$ and $\nabla\mathcal{L}_{ent}(\theta;\mathcal{X}^t_i)$, formulated as:
\begin{equation}
\begin{aligned}
\eta = \beta * Cus(\frac{\nabla\mathcal{L}_{con}(\theta^{'};\mathcal{X}^t_i) \cdot \nabla\mathcal{L}_{ent}(\theta;\mathcal{X}^t_i)}{\left \| \nabla\mathcal{L}_{con}(\theta^{'};\mathcal{X}^t_i) \right \| \left \| \nabla\mathcal{L}_{ent}(\theta;\mathcal{X}^t_i)\right \| }),
\end{aligned}
\label{lr}
\end{equation}
where $\beta$ is a scaling factor, and $Cus(x) = \frac{1}{4}(x + 1)^2$ is a custom increasing activation function to map the cosine similarity to $[0, 1]$.
As the cosine similarity increases, the optimization directions of $\nabla\mathcal{L}_{con}(\theta^{'};\mathcal{X}^t_i)$ and $\nabla\mathcal{L}_{ent}(\theta;\mathcal{X}^t_i)$ become more aligned, thus assigning a larger learning rate, and conversely.
The overall process of our GraTa is summarized in Algorithm~\ref{algorithm1}.


\begin{table*}[!htb]
    \centering
    \resizebox{0.88\textwidth}{!}{
    \begin{tabular}{c|c|ccccc|c}
        \Xhline{1pt}
        \multicolumn{2}{c|}{\multirow{2}*{Methods}} & 
        \multicolumn{1}{c}{Domain A} & 
        \multicolumn{1}{c}{Domain B} & 
        \multicolumn{1}{c}{Domain C} & 
        \multicolumn{1}{c}{Domain D} & 
        \multicolumn{1}{c|}{Domain E} & 
        Average \\ 
        
        \Xcline{3-8}{0.4pt}
        \multicolumn{1}{c}{} & &
        $DSC$ &
        $DSC$ &
        $DSC$ &
        $DSC$ &
        $DSC$ &
        $DSC\uparrow$ \\
        \hline 

         \multicolumn{2}{c|}{No Adapt} &
         $67.96$ & 
         $76.71$ & 
         $74.71$ & 
         $54.07$ &
         $68.88$ &
         $68.47$ \\
         \hline

         {\multirow{3}*{BN-based}} & DUA~\cite{DUA} &
         $73.10$ & 
         $77.10$ & 
         $75.34$ & 
         $58.62$ &
         $72.99$ &
         $71.43$ \\

         & DIGA~\cite{Dynamically} &
         $76.57$ & 
         $77.10$ & 
         $73.64$ & 
         $62.04$ & 
         $71.54$ & 
         $72.18$ \\ 

         & MedBN~\cite{MedBN} &
         $75.14$ & 
         $76.77$ & 
         $74.46$ & 
         $59.91$ & 
         $72.00$ & 
         $71.65$ \\ 
         \hline

         {\multirow{5}*{Fine-tune-based}} & TENT~\cite{TENT} &
         $74.06$ & 
         $78.55$ & 
         $74.56$ & 
         $51.45$ &
         $69.66$ &
         $69.65$ \\

         & DLTTA~\cite{DLTTA} &
         $75.19$ & 
         $78.48$ & 
         $76.37$ & 
         $57.50$ &
         $71.35$ &
         $71.78$ \\

         & DomainAdaptor~\cite{DomainAdaptor} &
         $75.88$ &
         $77.43$ &
         $75.66$ &
         $58.64$ &
         $72.38$ &
         $72.00$ \\ 

         & SAR~\cite{SAR} &
         $75.26$ & 
         $78.07$ & 
         $75.07$ & 
         $61.24$ &
         $\textbf{73.69}$ &
         $72.67$ \\

         & DeTTA~\cite{DeTTA} &
         $75.15$ & 
         $78.17$ & 
         $75.27$ & 
         $61.26$ & 
         $73.61$ & 
         $72.69$ \\ 
         \hline

         \multicolumn{2}{c|}{GraTa (Ours)} &
         $\textbf{76.61}$ & 
         $\textbf{78.81}$ & 
         $\textbf{76.52}$ & 
         $\textbf{66.84}$ &
         $72.94$ &
         $\textbf{74.34}$ \\
         \Xhline{1pt}

    \end{tabular}
    }
    \caption{Performance of our GraTa, `No Adapt' baseline, and eight TTA methods. The best results are highlighted in \textbf{bold}.}
    \label{tab:Comparison}
\end{table*}

\begin{table*}[!htb]
    \centering
    \resizebox{0.88\textwidth}{!}{
    \begin{tabular}{c|c|ccccc|c}
        \Xhline{1pt}
        {\multirow{2}*{Objective}} & 
        {\multirow{2}*{Learning Rate}} & 
        \multicolumn{1}{c}{Domain A} & 
        \multicolumn{1}{c}{Domain B} & 
        \multicolumn{1}{c}{Domain C} & 
        \multicolumn{1}{c}{Domain D} & 
        \multicolumn{1}{c|}{Domain E} & 
        Average \\ 
        
        \Xcline{3-8}{0.4pt}
         &
         &
        $DSC$ &
        $DSC$ &
        $DSC$ &
        $DSC$ &
        $DSC$ &
        $DSC\uparrow$ \\
        \hline

         - & - &
         $67.96$ & 
         $76.71$ & 
         $74.71$ & 
         $54.07$ &
         $68.88$ &
         $68.47$ \\
         \hline

        $\mathcal{L}_{con} (\theta;\mathcal{X}^t_i)$ & $\beta$ &
         $\textbf{78.35}$ & 
         $78.50$ & 
         $69.24$ & 
         $52.13$ & 
         $68.74$ & 
         $69.39$ \\
        \hline

          $\mathcal{L}_{con} (\theta-\nabla\mathcal{L}_{ent}(\theta;\mathcal{X}^t_i);\mathcal{X}^t_i)$ & $\beta$ &
         $77.97$ & 
         $78.77$ & 
         $73.90$ & 
         $56.53$ & 
         $70.46$ & 
         $71.53$  \\
        \hline

         $\mathcal{L}_{con} (\theta-\nabla\mathcal{L}_{ent}(\theta;\mathcal{X}^t_i);\mathcal{X}^t_i)$ & $\eta$ &
         $76.61$ & 
         $\textbf{78.81}$ & 
         $\textbf{76.52}$ & 
         $\textbf{66.84}$ &
         $\textbf{72.94}$ &
         $\textbf{74.34}$ \\
         \hline


        \Xhline{1pt}
    \end{tabular}
    }
    \caption{Performance of using various objectives and learning rates for optimization. The best results are highlighted in \textbf{bold}.}
    \label{tab:ablation}
\end{table*}

\section{Experiments and Results}
\subsection{Datasets and Evaluation Metrics}
We evaluate our proposed GraTa and other state-of-the-art TTA methods on the joint optic disc (OD) and cup (OC) segmentation task, which comprises five public datasets collected from different medical centres, denoted as domain A (RIM-ONE-r3~\cite{RIM_ONE_r3}), B (REFUGE~\cite{REFUGE}), C (ORIGA~\cite{ORIGA}), D (REFUGE-Validation/Test~\cite{REFUGE}), and E (Drishti-GS~\cite{Drishti-GS}).
These datasets consist of 159, 400, 650, 800, and 101 images, respectively.
For each image, we cropped a region of interest (ROI) centered at the OD with a size of $800\times 800$, and each ROI is further resized to $512\times 512$ and normalized by min-max normalization following~\cite{ProSFDA}.
We utilize the Dice score metric ($DSC$) for evaluation.

\subsection{Implementation Details}
We trained a ResUNet-34~\cite{ResNet} backbone following~\cite{ProSFDA} as the baseline individually on each domain (source domain) and subsequently tested it on each remaining domain (target domain), computing the mean metrics to evaluate all the methods across diverse scenarios ($4 \times 5$ in total).
For a fair comparison, we conducted single-iteration adaptation for each batch of test data using a batch size of 1 across all experiments following~\cite{DLTTA}.
To deploy our GraTa, we utilized the Adam~\cite{Adam} optimizer to train the affine-transformation parameters within pre-trained models, and the BN statistics were recollected from test data following~\cite{TENT}.
The scaling factor $\beta$ is set to 0.0001 empirically.

\begin{figure*}[!htb]
    \centering
    \includegraphics[width=0.95\textwidth]{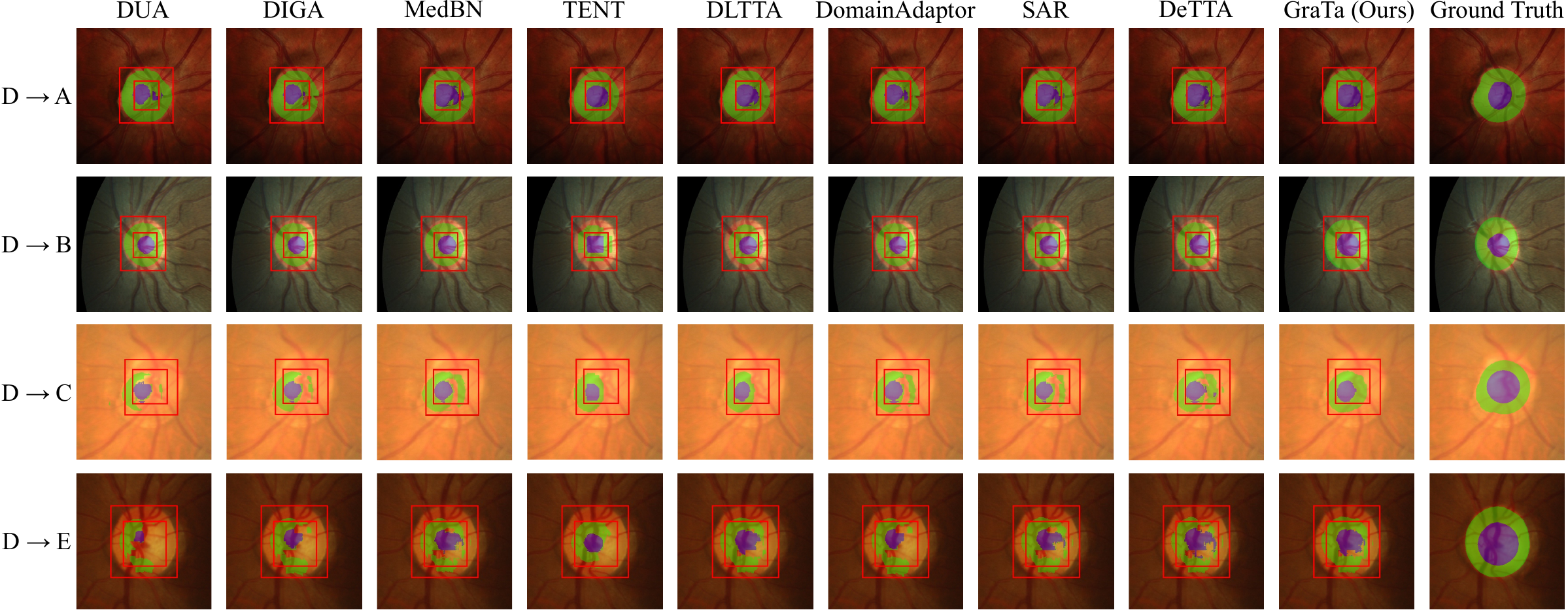}
    \caption{Qualitative results of our GraTa and eight TTA methods using Domain D as the source domain and remaining domains as target domains. We displayed the results of two simple samples (rows 1-2) and two hard samples (rows 3-4). The ground-truth bounding boxes of OD and OC were overlaid on each segmentation result to highlight potential over- or under-segmentation. Best viewed in color.
}
	\label{fig-seg}
\end{figure*}

\subsection{Results}
We compare our GraTa with the `No Adapt' baseline (testing without adaptation), and eight competing TTA methods, including three methods based on refining the BN statistics (DUA~\cite{DUA}, DIGA~\cite{Dynamically}, and MedBN~\cite{MedBN}), and five methods based on fine-tuning the pre-trained models (TENT~\cite{TENT}, DLTTA~\cite{DLTTA}, DomainAdaptor~\cite{DomainAdaptor}, SAR~\cite{SAR}, and DeTTA~\cite{DeTTA}).
Specifically, we re-implemented all competing methods using the same baseline as our GraTa and integrated DLTTA with the entropy loss proposed in TENT~\cite{DLTTA}. 
The result of each domain is the average metric calculated by treating the current domain as the source domain and employing the pre-trained model on remaining domains.

\begin{table*}[!htb]
    \centering
    \resizebox{0.88\textwidth}{!}{
    \begin{tabular}{c|c|ccccc|c}
        \Xhline{1pt}
        {\multirow{2}*{Optimization Objective}} & 
        {\multirow{2}*{Learning Rate}} & 
        \multicolumn{1}{c}{Domain A} & 
        \multicolumn{1}{c}{Domain B} & 
        \multicolumn{1}{c}{Domain C} & 
        \multicolumn{1}{c}{Domain D} & 
        \multicolumn{1}{c|}{Domain E} & 
        Average \\ 
        
        \Xcline{3-8}{0.4pt}
         &
         &
        $DSC$ &
        $DSC$ &
        $DSC$ &
        $DSC$ &
        $DSC$ &
        $DSC\uparrow$ \\
        \hline

         $\mathcal{L}_{ent} (\theta;\mathcal{X}^t_i)$ & $\beta$ &
         $74.06$ & 
         $78.55$ & 
         $74.56$ & 
         $51.45$ &
         $69.66$ &
         $69.65$ \\
         \hline

         $\mathcal{L}_{ent} (\theta-\nabla\mathcal{L}_{con}(\theta;\mathcal{X}^t_i);\mathcal{X}^t_i)$ & $\eta$ &
         $74.69$ & 
         $78.46$ & 
         $75.60$ & 
         $56.75$ &
         $71.86$ &
         $71.47$ \\
         \hline
         
         $\mathcal{L}_{con} (\theta-\nabla\mathcal{L}_{ent}(\theta;\mathcal{X}^t_i);\mathcal{X}^t_i)$ & $\eta$ &
         $\textbf{76.61}$ & 
         $\textbf{78.81}$ & 
         $\textbf{76.52}$ & 
         $\textbf{66.84}$ &
         $\textbf{72.94}$ &
         $\textbf{74.34}$ \\
        \Xhline{1pt}
    \end{tabular}
    }
    \caption{Performance of TENT, our GraTa, and its variant. The best results are highlighted in \textbf{bold}.}
    \label{tab:exchange}
\end{table*}

\noindent{\textbf{Comparing to other TTA methods.}}
The results of our GraTa, the “No Adapt” baseline, and eight competing TTA methods are detailed in Table~\ref{tab:Comparison}. 
The BN-based methods exhibit stable but suboptimal enhancements across all domains due to the absence of model updating.
TENT exhibits limited performance, while other variants (\emph{i.e.}, DLTTA, DomainAdaptor, and SAR) enhance the performance significantly.
It can be observed that our GraTa achieves the best performance across most scenarios and improves the baseline by $5.87\%$, which demonstrates the effectiveness and superiority of our GraTa.
Specifically, our GraTa boosts the performance in Domain D by $12.77\%$, while other methods exhibit marginal improvement, which means our approach extends the upper limit of TTA in challenging scenarios.

\begin{table}[!htb]
    \centering
    \resizebox{\columnwidth}{!}{
    \begin{tabular}{c|ccccc|c}
        \Xhline{1pt}
        {\multirow{2}*{Function}} & 
        \multicolumn{1}{c}{Do. A} & 
        \multicolumn{1}{c}{Do. B} & 
        \multicolumn{1}{c}{Do.n C} & 
        \multicolumn{1}{c}{Do. D} & 
        \multicolumn{1}{c|}{Do. E} & 
        Average \\ 
        
        \Xcline{2-7}{0.4pt}
         &
        $DSC$ &
        $DSC$ &
        $DSC$ &
        $DSC$ &
        $DSC$ &
        $DSC\uparrow$ \\
        \hline

         Linear: $\frac{1}{2}(x + 1)$ &
         $77.15$ & 
         $\textbf{79.11}$ & 
         $76.07$ & 
         $63.79$ &
         $72.42$ &
         $73.71$ \\
         \hline

         Sigmoid: $\frac{1}{1+e^{-x}}$ &
         $77.31$ & 
         $79.04$ & 
         $76.01$ & 
         $64.69$ &
         $72.10$ &
         $73.83$ \\
         \hline
 	 	 	 		 
         ReLU: $max(0, x)$ &
         $75.92$ & 
         $78.29$ & 
         $75.16$ & 
         $62.14$ &
         $\textbf{73.07}$ &
         $72.92$ \\
         \hline
 	 	 	 
         Softplus: $\ln{(1+e^x)}$ &
         $\textbf{77.60}$ & 
         $78.81$ & 
         $75.44$ & 
         $63.04$ &
         $71.68$ &
         $73.32$ \\
         \hline

         Ours: $\frac{1}{4}(x + 1)^2$ &
         $76.61$ & 
         $78.81$ & 
         $\textbf{76.52}$ & 
         $\textbf{66.84}$ &
         $72.94$ &
         $\textbf{74.34}$ \\
        \Xhline{1pt}
    \end{tabular}
    }
    \caption{Performance of using different activation functions to map the cosine similarity. $e$ denotes the natural base. The best results are highlighted in \textbf{bold}.}
    \label{tab:function}
\end{table}

\noindent{\textbf{Ablation study.}}
To evaluate the contributions of our optimization objective and dynamic learning rate, we conducted a series of ablation experiments, as shown in Table~\ref{tab:ablation}. 
It shows that (1) using $\mathcal{L}_{con} (\theta;\mathcal{X}^t_i)$ only obtains limited performance similar to TENT; (2) introducing $\mathcal{L}_{con} (\theta-\nabla\mathcal{L}_{ent}(\theta;\mathcal{X}^t_i);\mathcal{X}^t_i)$ and $\eta$ enhances the performance by $2.14\%$ and $2.81\%$, respectively; (3) the best performance is achieved when both are utilized jointly (\emph{i.e.}, our GraTa).

\noindent{\textbf{Qualitative analysis.}}
We selected the model trained on Domain D (source) as a case study and visualized four test images from four scenarios, denoted as ``D $\rightarrow$ target domain,'' along with their corresponding segmentation results produced by our GraTa method, eight competing methods, and the ground truth, as shown in Figure~\ref{fig-seg}. The regions of OD and OC are highlighted in green and blue, respectively. For comparative purposes, we drew the bounding boxes of the OD and OC on the ground truth and overlaid them on each segmentation result. Compared to other competing methods, our GraTa demonstrates reduced over-segmentation and under-segmentation across all samples, indicating the effectiveness of the improved optimization procedure.

\subsection{Discussions}

\noindent{\textbf{Exchange the Location of $\mathcal{L}_{con}$ and $\mathcal{L}_{ent}$.}}
It is particularly critical to determine which objective function is the optimal one to produce the pseudo gradient.
We exchanged the location of $\mathcal{L}_{con}$ and $\mathcal{L}_{ent}$ in our objective and repeated the experiments for evaluation.
The results are listed in Table~\ref{tab:exchange}.
Compared to TENT which optimizes $\mathcal{L}_{ent} (\theta;\mathcal{X}^t_i)$ only, the variant of our GraTa using $\mathcal{L}_{ent} (\theta-\nabla\mathcal{L}_{con}(\theta;\mathcal{X}^t_i);\mathcal{X}^t_i)$ as the optimization objective and $\eta$ as the learning rate also boosts the performance.
Our GraTa achieves the best performance, which demonstrates that $\mathcal{L}_{con}$ is more suitable to produce the pseudo gradient due to its superior supervision.

\noindent{\textbf{Obtain $\eta$ using other activation functions.}}
To ascertain the optimal activation function for mapping the cosine similarity to obtain $\eta$, we conducted experiments using various activation functions and listed the results in Table~\ref{tab:function}. It reveals that (1) our designed function achieves superior performance to others by amplifying the distinctions among learning rates generated by diverse cosine similarity scores; (2) ReLU exhibits the worst average performance which underscores the necessity of assigning appropriate learning rates for fine-tuning even if distinct gradients diverge largely.

\begin{figure*}[!htb]
    \centering
    \includegraphics[width=0.65\textwidth]{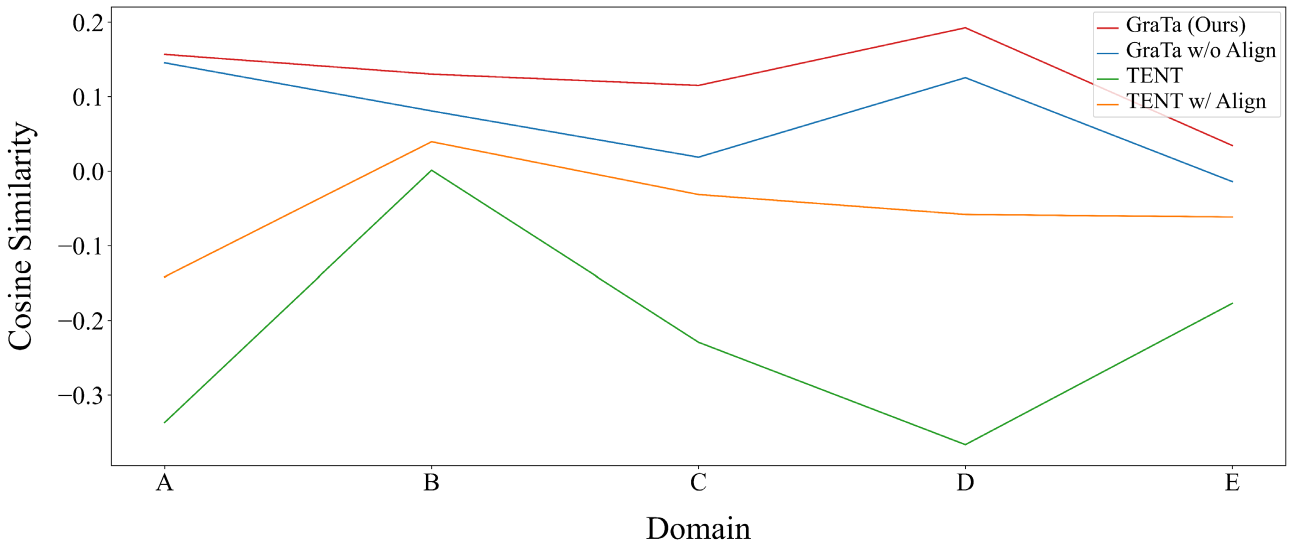}
    \caption{Cosine similarity between the pseudo gradient and empirical gradient w/ or w/o our proposed gradient alignment. The objectives of ``TENT'' and ``TENT w/ Align'' are $\mathcal{L}_{ent} (\theta;\mathcal{X}^t_i)$ and $\mathcal{L}_{ent} (\theta-\nabla\mathcal{L}_{con} (\theta;\mathcal{X}^t_i);\mathcal{X}^t_i)$, respectively.}
    \label{fig-cosine}
\end{figure*}

\noindent{\textbf{Alignment between the empirical and pseudo gradients.}}
To demonstrate that GraTa can align the empirical gradient and pseudo gradient more closely, we repeated the experiments on all scenarios and visualized the cosine similarity between these two gradients, where the empirical gradient is produced by the cross-entropy loss in a supervised manner, as shown in Figure~\ref{fig-cosine}.
We reported the average results using each domain as the source domain and remaining ones as target domains.
The greater the cosine similarity, the closer the pseudo gradient aligns with the empirical gradient.
The results reveal the effectiveness of our GraTa in improving gradient direction and also indicate that even small alignment improvements can lead to significant performance gains.

\begin{table}[!htb]
    \centering
    \resizebox{\columnwidth}{!}{
    \begin{tabular}{c|c|ccccc|c}
        \Xhline{1pt}
        {\multirow{2}*{Aux}} & 
        {\multirow{2}*{Pse}} & 
        \multicolumn{1}{c}{Do. A} & 
        \multicolumn{1}{c}{Do. B} & 
        \multicolumn{1}{c}{Do. C} & 
        \multicolumn{1}{c}{Do. D} & 
        \multicolumn{1}{c|}{Do. E} & 
        Average \\ 
        
        \Xcline{3-8}{0.4pt}
         &
         &
        $DSC$ &
        $DSC$ &
        $DSC$ &
        $DSC$ &
        $DSC$ &
        $DSC\uparrow$ \\
        \hline

         $\mathcal{L}_{ent}$ &  {\multirow{5}*{$\mathcal{L}_{con}$}} &
         $76.61$ & $78.81$ & $76.52$ & $66.84$ & $72.94$ & $\textbf{74.34}$ \\
         $\mathcal{L}_{rec}$ &   &
         $\textbf{76.76}$ & $78.94$ & $74.74$ & $66.43$ & $72.19$ & $73.81$ \\
         $\mathcal{L}_{rot}$ &   &
         $76.58$ & $\textbf{79.02}$ & $74.39$ & $\textbf{67.90}$ & $72.53$ & $74.08$ \\
         $\mathcal{L}_{sup}$ &   &
         $76.68$ & $78.99$ & $74.32$ & $67.82$ & $72.31$ & $74.02$ \\ 
         $\mathcal{L}_{den}$ &   &
         $76.71$ & $78.94$ & $75.11$ & $65.18$ & $72.26$ & $73.64$ \\
         \hline
         
         $\mathcal{L}_{con}$ &  {\multirow{5}*{$\mathcal{L}_{ent}$}} &
         $74.69$ & $78.46$ & $75.60$ & $56.75$ & $71.86$ & $71.47$ \\
         $\mathcal{L}_{rec}$ &   &
         $75.27$ & $78.54$ & $76.36$ & $60.29$ & $70.97$ & $72.28$ \\
         $\mathcal{L}_{rot}$ &   &
         $75.19$ & $78.54$ & $76.34$ & $56.29$ & $70.78$ & $71.43$ \\
         $\mathcal{L}_{sup}$ &   &
         $75.20$ & $78.53$ & $76.53$ & $57.41$ & $70.79$ & $71.69$ \\
         $\mathcal{L}_{den}$ &   &
         $75.37$ & $78.52$ & $\textbf{76.67}$ & $59.38$ & $70.72$ & $72.13$ \\
         \hline

         $\mathcal{L}_{con}$ &  {\multirow{5}*{$\mathcal{L}_{rec}$}} &
         $75.18$ & $77.85$ & $75.13$ & $62.33$ & $73.71$ & $72.84$ \\
         $\mathcal{L}_{ent}$ &   &
         $74.47$ & $77.74$ & $74.71$ & $59.64$ & $73.49$ & $72.01$ \\
         $\mathcal{L}_{rot}$ &   &
         $75.74$ & $78.04$ & $75.12$ & $60.13$ & $\textbf{73.91}$ & $72.59$ \\
         $\mathcal{L}_{sup}$ &   &
         $74.58$ & $77.51$ & $74.91$ & $62.99$ & $73.67$ & $72.73$ \\ 
         $\mathcal{L}_{den}$ &   &
         $75.44$ & $77.82$ & $75.66$ & $60.12$ & $73.73$ & $72.55$ \\
         \hline

         $\mathcal{L}_{con}$ &  {\multirow{5}*{$\mathcal{L}_{rot}$}} &
         $75.31$ & $78.01$ & $75.02$ & $61.66$ & $73.81$ & $72.76$ \\
         $\mathcal{L}_{ent}$ &   &
         $75.37$ & $78.17$ & $75.07$ & $62.14$ & $73.75$ & $72.90$ \\
         $\mathcal{L}_{rec}$ &   &
         $75.19$ & $78.10$ & $74.39$ & $61.57$ & $73.78$ & $72.60$ \\ 
         $\mathcal{L}_{sup}$ &   &
         $75.31$ & $78.08$ & $75.95$ & $61.24$ & $73.73$ & $72.86$ \\ 
         $\mathcal{L}_{den}$ &   &
         $75.27$ & $78.21$ & $74.84$ & $61.56$ & $73.65$ & $72.70$ \\ 
         \hline

         $\mathcal{L}_{con}$ &  {\multirow{5}*{$\mathcal{L}_{sup}$}} &
         $75.36$ & $78.17$ & $75.27$ & $61.49$ & $73.77$ & $72.81$ \\ 
         $\mathcal{L}_{ent}$ &   &
         $75.51$ & $77.86$ & $74.67$ & $59.23$ & $73.47$ & $72.15$ \\ 
         $\mathcal{L}_{rec}$ &   &
         $75.26$ & $78.09$ & $74.53$ & $59.78$ & $73.66$ & $72.27$ \\ 
         $\mathcal{L}_{rot}$ &   &
         $75.17$ & $77.87$ & $74.92$ & $58.48$ & $73.75$ & $72.04$ \\ 
         $\mathcal{L}_{den}$ &   &
         $74.94$ & $77.91$ & $74.83$ & $60.65$ & $73.63$ & $72.39$ \\ 
         \hline

         $\mathcal{L}_{con}$ &  {\multirow{5}*{$\mathcal{L}_{den}$}} &
         $75.34$ & $78.02$ & $74.47$ & $61.74$ & $73.71$ & $72.66$ \\
         $\mathcal{L}_{ent}$ &   &
         $75.20$ & $77.44$ & $74.77$ & $59.22$ & $73.38$ & $72.00$ \\ 
         $\mathcal{L}_{rec}$ &   &
         $75.40$ & $78.09$ & $74.51$ & $60.71$ & $73.59$ & $72.46$ \\
         $\mathcal{L}_{rot}$ &   &
         $75.32$ & $77.71$ & $76.49$ & $55.30$ & $73.44$ & $71.65$ \\ 
         $\mathcal{L}_{sup}$ &   &
         $75.48$ & $77.87$ & $75.08$ & $59.82$ & $73.68$ & $72.38$ \\ 
         \hline
        \Xhline{1pt}
    \end{tabular}
    }
    \caption{Performance of various optimization objective combinations, where $\mathcal{L}_{con}$, $\mathcal{L}_{ent}$, $\mathcal{L}_{rec}$, $\mathcal{L}_{rot}$, $\mathcal{L}_{sup}$, $\mathcal{L}_{den}$ denote the consistency loss, the entropy loss, the reconstruction loss, the rotation prediction loss, the super-resolution loss, and the denoising loss, respectively. The best results are highlighted in \textbf{bold}. `Aux': Abbreviation of `Auxiliary'. `Pse': Abbreviation of `Pseudo'.}
    \label{tab:combination}
\end{table}

\noindent{\textbf{More combinations of objective functions.}}
For further discussion, we introduced four additional loss functions, \emph{i.e.}, the reconstruction loss, the rotation prediction loss, the super-resolution loss, and the denoising loss. 
We utilized the cross-entropy loss to calculate the rotation prediction loss and the mean-square-error loss to calculate others.
To construct the super-resolution loss, we downsampled the input image by a factor of four. 
For the denoising loss, we added the Gaussian noise sampled from a standard normal distribution to the image.
We evaluated all possible combinations of them to determine the optimal one and displayed the results in Table~\ref{tab:combination}.
It reveals that (1) using $\mathcal{L}_{con}$ as the pseudo objective function achieves superior performance over others; (2) using $\mathcal{L}_{rec}$, $\mathcal{L}_{rot}$, $\mathcal{L}_{sup}$, or $\mathcal{L}_{den}$ as the pseudo objective function results in similar but suboptimal performance since they all require an additional branch to construct the loss, making it challenging to provide appropriate gradients during inference; (3) using $\mathcal{L}_{rec}$, $\mathcal{L}_{rot}$, $\mathcal{L}_{sup}$, or $\mathcal{L}_{den}$ as the auxiliary objective function can enhance the performance of $\mathcal{L}_{con}$ and $\mathcal{L}_{ent}$, demonstrating the effectiveness of GraTa.


\section{Conclusion}
In this paper, we proposed the GraTa, a TTA method to focus on optimizing the optimization direction and step-size during inference by gradient alignment and utilizing a dynamic learning rate, respectively.
Specifically, we introduced an auxiliary gradient to the pseudo one to perform alignment to excavate the similarities between distinct gradients and further approximate the empirical gradient.
The dynamic learning rate is constructed based on the cosine similarity between the pseudo and auxiliary gradients, encouraging pre-trained models to enhance learning as these gradients become closely aligned.
Extensive experiment results on diverse scenarios demonstrated the superiority of our GraTa and the effectiveness of each special design.
In our future work, we intend to explore the alignment of other auxiliary gradients to further enhance performance, as well as investigate other deployment scenarios, such as continual test-time adaptation~\cite{chen2024each}.
We hope that our efforts can promote research on gradient optimization for TTA.

\section{Acknowledgments}
This work was supported in part by the National Natural Science Foundation of China (Nos. 62171377, 92470101, 6240012686), in part by grants from National Key R\&D Program of China (2022YFC2009903/2022YFC2009900), in part by the Ningbo Clinical Research Center for Medical Imaging under Grant 2021L003 (Open Project 2022LYKFZD06), in part by Shenzhen Science and Technology Program under Grants JCYJ20220530161616036, in part by the Fundamental Research Funds for the Central Universities (No. D5000230376), and in part by the Innovation Foundation for Doctor Dissertation of Northwestern Polytechnical University under Grant CX2024016.

\bibliography{aaai25}

\end{document}